\documentclass[conference]{IEEEtran}
\IEEEoverridecommandlockouts
\usepackage{amsmath,amssymb,amsfonts}
\usepackage{algorithmic}
\usepackage{graphicx}
\usepackage{textcomp}
\usepackage{xcolor}
\usepackage{array}
\usepackage{mdwmath}
\usepackage{mdwtab}
\usepackage{eqparbox}
\usepackage{url}
\usepackage{hyperref}
\usepackage{tabularx}
\usepackage{subcaption}
\hypersetup{citecolor=Green}
\usepackage{mathrsfs}
\usepackage{rsfso}
\usepackage{multirow}
\usepackage{amssymb}
\usepackage{lipsum}
\usepackage{subfloat}
\usepackage[style=ieee, sorting=none]{biblatex}
\usepackage{comment}
\def\BibTeX{{\rm B\kern-.05em{\sc i\kern-.025em b}\kern-.08em
    T\kern-.1667em\lower.7ex\hbox{E}\kern-.125emX}}

\makeatletter
 \let\old@ps@headings\ps@headings
 \let\old@ps@IEEEtitlepagestyle\ps@IEEEtitlepagestyle
 \def\confheader#1{%
 
 \def\ps@IEEEtitlepagestyle{%
 \old@ps@IEEEtitlepagestyle%
 \def\@oddhead{\strut\hfill#1\hfill\strut}%
 \def\@evenhead{\strut\hfill#1\hfill\strut}%
 }%
 \ps@headings%
 }
 \makeatother

\begin{document}

\title{Comparative Evaluation of Weather Forecasting
using Machine Learning Models
}

\author{
\IEEEauthorblockN{ Md Saydur Rahman\IEEEauthorrefmark{1},Farhana Akter Tumpa\IEEEauthorrefmark{2},Md Shazid Islam\IEEEauthorrefmark{1},\\ Abul Al Arabi\IEEEauthorrefmark{3}, Md Sanzid Bin Hossain\IEEEauthorrefmark{4}, Md Saad Ul Haque\IEEEauthorrefmark{5}}

\IEEEauthorblockA{\IEEEauthorrefmark{1}University of California Riverside, USA,\\ \IEEEauthorrefmark{2}Ahsanullah University of Science and Technology, Bangladesh \\ \IEEEauthorrefmark{3}CSE, Texas A\&M University, USA\\\IEEEauthorrefmark{4}University of Central Florida, USA,
\IEEEauthorrefmark{5} University of Florida, USA\\
Email:  mrahm054@ucr.edu, Tumpafarhanaakter@gmail.com,misla048@ucr.edu,\\abulalarabi@tamu.edu,
md543636@ucf.edu,
haque.m@ufl.edu,
}}


\maketitle

\begin{abstract}

Gaining a deeper understanding of weather and being able to predict its future conducts have always been considered important endeavors for the growth of our society. This research paper explores the advancements in understanding and predicting nature's behavior, particularly in the context of weather forecasting, through the application of machine learning algorithms. By leveraging the power of machine learning, data mining, and data analysis techniques, significant progress has been made in this field. This study focuses on analyzing the contributions of various machine learning algorithms in predicting precipitation and temperature patterns using a 20-year dataset from a single weather station in Dhaka city. Algorithms such as Gradient Boosting, AdaBoosting, Artificial Neural Network, Stacking Random Forest, Stacking Neural Network, and Stacking KNN are evaluated and compared based on their performance metrics, including Confusion matrix measurements. The findings highlight remarkable achievements and provide valuable insights into their performances and features correlation.
\end{abstract}


\section{Introduction}
The influence of weather on human survival and the development of civilization has long been recognized as pivotal. Accurate prediction of precipitation and temperature assumes paramount importance for the assessment of hydrological processes, effective management of water resources, anticipation of floods and droughts, and the evaluation of climate change impacts. Furthermore, the escalating rate of global warming and climate-related disasters has heightened the urgency for weather data prediction. While rain gauges provide localized data, radar-based models offer enhanced spatiotemporal resolution. Approaches to precipitation forecasting encompass dynamic methods based on physical models as well as empirical techniques reliant on data-driven methodologies. Notably, the empirical approach, leveraging artificial intelligence algorithms, has demonstrated promise in tackling meteorological variables' intricate and chaotic nature, thereby facilitating efficient and reliable predictions. Machine learning models, widely explored in the existing body of literature, have gained substantial recognition as effective tools for weather forecasting.\\
Machine learning algorithms have been very useful in processing multi-modal data \cite{b19,b21,b9,power3,power4,power5,araf} and optimization \cite{power1,power2}. These algorithms
have been extensively researched and compared to assess their accuracy in predicting continually changing phenomena such as weather conditions. Key factors such as temperature, humidity, wind speed, surface pressure, and precipitation are vital in obtaining precise measurements for weather pattern predictions. Various popular machine learning algorithms, including Artificial Neural Networks (ANN), Support Vector Machines (SVM), Support Vector Regression (SVR), and Genetic Algorithms (GA), have been utilized with specific modifications or adaptations in these techniques. Recent advancements in Numerical Weather Prediction (NWP) models have enabled testing at finer resolutions, facilitating a more detailed representation of storm processes and hourly predictions. However, effectively placing and timing precipitation and temperature remains a challenge compared to models with coarser resolutions. \\
In order to tackle this challenge, ensemble methodologies utilizing storm-scale Numerical Weather Prediction (NWP) models have emerged, offering enhanced estimations of uncertainty by effectively sampling spatiotemporal errors linked to individual storms. The integration of ensemble predictions into accurate and actionable consensus guidance presents an area warranting further investigation. Boosting algorithms, such as AdaBoost, XGBoost, and Gradient Boost, have garnered considerable attention within the field of weather prediction, as they iteratively combine weak rules to establish a robust learner, thereby yielding more precise outcomes. Extensive analyses have been conducted to evaluate the performance of these algorithms in discerning weather conditions based on multiple contributing factors.\\
Bangladesh experiences significant fluctuations in rainfall and temperature, particularly in its capital city, Dhaka, thereby necessitating targeted investigations. The present study is dedicated to the analysis of an extensive collection of weather data encompassing a time span of around two decades (2003-2023) obtained from Dhaka City, Bangladesh. The dataset comprises a diverse array of crucial variables, including wind speed, maximum and minimum temperatures, precipitation, wind direction, surface pressure, relative humidity, and specific humidity. In order to extract meaningful insights from the dataset, we have employed a suite of sophisticated algorithms, namely Gradient Boost, AdaBoost, Stacking KNN, and Stacking Random Forest. Our primary objective revolves around assessing the accuracy and performance of these algorithms in predicting long-term precipitation patterns and average temperature. Through the utilization of these advanced methodologies, we aim to enrich our comprehension of weather dynamics and bolster our forecasting capabilities in the Dhaka region.

\section{Related Works}

 There has been a lot of work progressed in the context of weather pattern forecasting by machine learning. Recent research on IoT-based environment data collection (e.g.,\cite{iot}) is further fostering the weather prediction paradigm. In weather forecast research, authors involve the creation of a dataset by collecting data from previous months \cite{b7}. To simplify the analysis, the months are categorized into three seasons: summer, fall, and winter, with each season comprising four months. Subsequently, algorithms like linear regression, logistic regression, and Gaussian Naive Bayes are applied to the dataset for further analysis and prediction purposes.\\
In \cite{b9} authors aim to develop precipitation forecasting models for Northern Bangladesh with a lag time of up to 3 months. Two machine learning algorithms, M5P and SVR, were used, and a hybrid model combining both (M5P-SVR) was developed. The SVR parameters were optimized using the particle swarm optimization algorithm. This hybrid model is unique in the literature for predicting monthly precipitation in the monsoon climate of Northern Bangladesh. The study also includes a sensitivity analysis of the number of input parameters and lag time.\\
Moreover, the author proposes a data augmentation algorithm combining K-means clustering and SMOTE to enhance sample information for forecasting monthly grid precipitation in the Danjiangkou River Basin\cite{b10}. RF, XGB, RNN, and LSTM models are employed. The aim is to improve medium- and long-term precipitation forecasting accuracy. Results show that SMOTE-km-XGB performs best, and deep learning methods (RNN and LSTM) do not benefit significantly from the data augmentation. This study enhances precipitation forecast accuracy and provides insights for improving hydrological forecasting.\\
Here, authors analyze 20 years of weather data from Chittagong city in Bangladesh, including variables such as wind speed, temperature, precipitation, and surface pressure\cite{b11}. Various algorithms, including AdaBoost, XGBoost, Stacking KNN, and Stacking Random Forest, were applied to predict maximum temperature, minimum temperature, and average temperature over a long-term duration. The accuracy and performance of these algorithms were compared and analyzed in this study.\\
Utility companies facing a lack of access to a weather service that provides predicted hourly temperatures for short-term load forecasting is indeed a significant concern. To overcome this limitation, a temperature forecaster was developed, capable of forecasting hourly temperatures up to seven days in advance \cite{b12}. The mean absolute inaccuracy of the one-day predictions by this forecaster was found to be 1.48°F. The authors employed an Artificial Neural Network (ANN) for creating temperature predictions, allowing for adjustments in response to unusual meteorological events. However, it is important to note that training the ANN requires more computational resources compared to traditional forecasting methods.\\
In a study conducted in 2015, eight unique regression tree structures were employed for short-term wind speed prediction \cite{b13}. The author further compared the performance of the most effective regression tree approach against other artificial intelligence (AI) methods, including support vector regression (SVR), multi-layer perceptron (MLP), extreme learning machines, and multi-linear regression. The findings revealed that the best regression tree approach outperformed the other AI methods in wind speed prediction.\\
In \cite{b15}, a lightweight data-driven weather forecasting model has been proposed using LSTM and TCN approaches. It compares the model's performance with classical machine learning and statistical forecasting approaches, as well as the NWP model. The LSTM and TCN models capture weather information from time-series data and are evaluated in multi-input multi-output and multi-input single-output regressions. The proposed deep learning networks show promise for accurate weather prediction using surface weather parameters.\\
In \cite{power4}, authors examines a novel deep learning-based
precipitation prediction approach. 
their model can adapt to modify
in weather features such as temperature, wind flow, humidity,
and so on. They aim to deal the limitations of
traditional prediction models that only rely on target domain data by leveraging the information gained from
various source domains. By adapting the prediction models to the target domain, they show upgrade the accuracy of
rain precipitation prediction for multiple locations with various weather patterns.

\section{Methodology}

Data cleaning, or data mining, is a crucial process for extracting valuable insights from raw data and preparing it for input into mathematical models. We collected weather data from NASA Power and conducted an extensive analysis of 20 years' worth of Dhaka city weather data, spanning from January 1, 2003, to January 1, 2023 \cite{b20}. The location of the dataset is 23.8103° N and 90.4125° E.  The dataset was carefully segmented, focusing on specific target values related to temperature and precipitation. With 16 extracted weather features, we aimed to predict average temperatures and precipitation accurately. 
\subsection{Dataset}
Our findings from this dataset provide valuable insights into the relationships between different weather variables and offer valuable information for weather forecasting and understanding weather patterns in the Dhaka region. The extracted features are given below:


\begin{table}[htbp]
\caption{Features Used in the Dataset
}
\begin{center}
$\begin{array}{|c|c|}
\hline \textbf{ Features } & \begin{array}{c}
\textbf{ Description } 
\end{array} \\
\hline \begin{array}{c}
\text { T2M  } \\
\end{array} & \text{Average air temperature at 2 meters}  \\
\hline \begin{array}{c}
\text { T2MDEW} \\
\end{array} & \text{Dew point at 2 meters}\\
\hline \begin{array}{c}
\text {  T2MWET} \\
\end{array}& \text{Wet point at 2 meters}  \\
\hline \begin{array}{c}
\text { TS } \\
\end{array}& \text{Earth Skin Temperature} \\
\hline \begin{array}{c}
\text {T2M RANGE} \\

\end{array} & \text{ Merra-2 Temperature at 2 meters range} \\
\hline 
\begin{array}{c}
\text {  T2M MAX } \\
\end{array}& \text{Maximum temperature at 2 meters} \\
\hline 
\begin{array}{c}
\text {  T2M MIN } \\
\end{array}& \text{Minimum temperature at 2 meters} \\
\hline 
\begin{array}{c}
\text {  QV2M } \\
\end{array} &\text{Specific Humidity at 2 meters} \\
\hline 
\begin{array}{c}
\text {  RH2M } \\
\end{array}& \text{Relative Humidity at 2 meters} \\
\hline 
\begin{array}{c}
\text { PRECTOT } \\
\end{array} &\text{Bias-corrected total
precipitation at 2 meters} \\
\hline 
\begin{array}{c}
\text {  PS } \\
\end{array} &\text{Average surface pressure at surface} \\
\hline 
\begin{array}{c}
\text {  WS10M RANGE} \\
\end{array} &\text{Average wind speed 10 meters range} \\
\hline 
\begin{array}{c}
\text { WS10M } \\
\end{array} &\text{Average wind speed 10 meters high} \\
\hline 
\begin{array}{c}
\text {  WD10M } \\
\end{array} &\text{Average wind direction 10 meters high} \\
\hline 
\begin{array}{c}
\text {  WS10M MAX } \\
\end{array} &\text{maximum wind speed 10 meters} \\
\hline 
\begin{array}{c}
\text {  WS10M MIN } \\
\end{array} & \text{Minimum wind speed 10 meters}\\
\hline 

\end{array}$\\
\end{center}
\end{table}

\subsection{Data Visualisation and Analysis}

From the dataset, we have analyzed the correlation of data for precipitation and temperature. From fig-1 and fig-2 we have deciphered the density of the distribution for both of them. We have used MinMaxScaler to scale the dataset and standardize the distribution. From fig-3 heatmap, we have found the correlation between the features related to precipitation and temperature. Among the 16 features precipitation mostly correlated with dew point, specific humidity, and relative humidity whereas temperature mostly correlated with wet point and specific humidity. Figure 4 and 5 depicts the distribution of monthly precipitation and temperature.
We can find that in the middle of the year, both precipitation and temperature are higher than at the end and start of the year. 

\begin{figure}[htbp]
\centerline{\includegraphics[width=7cm, height=6.5cm]{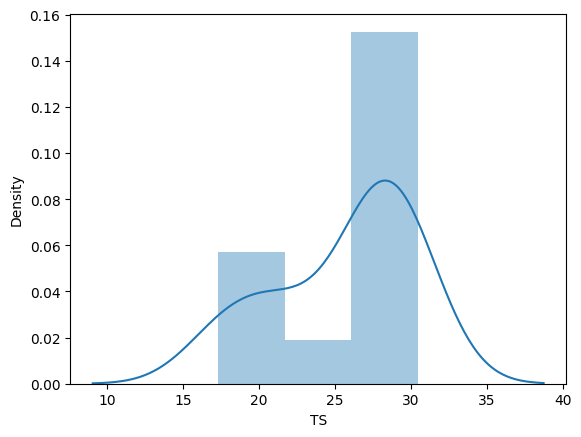}}
\caption{Distribution plot for Temperature}
\label{fig}
\end{figure}

\begin{figure}[htbp]
\centerline{\includegraphics[width=7cm, height=5cm]{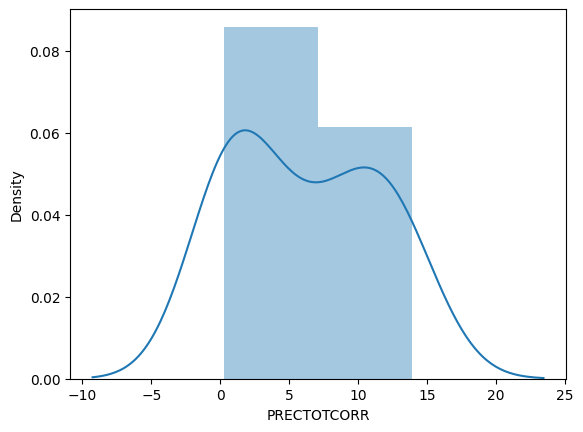}}
\caption{Distribution plot for Precipitation }
\label{fig}
\end{figure}

\begin{figure}[htbp]
\centerline{\includegraphics[width=8cm, height=7.5cm]{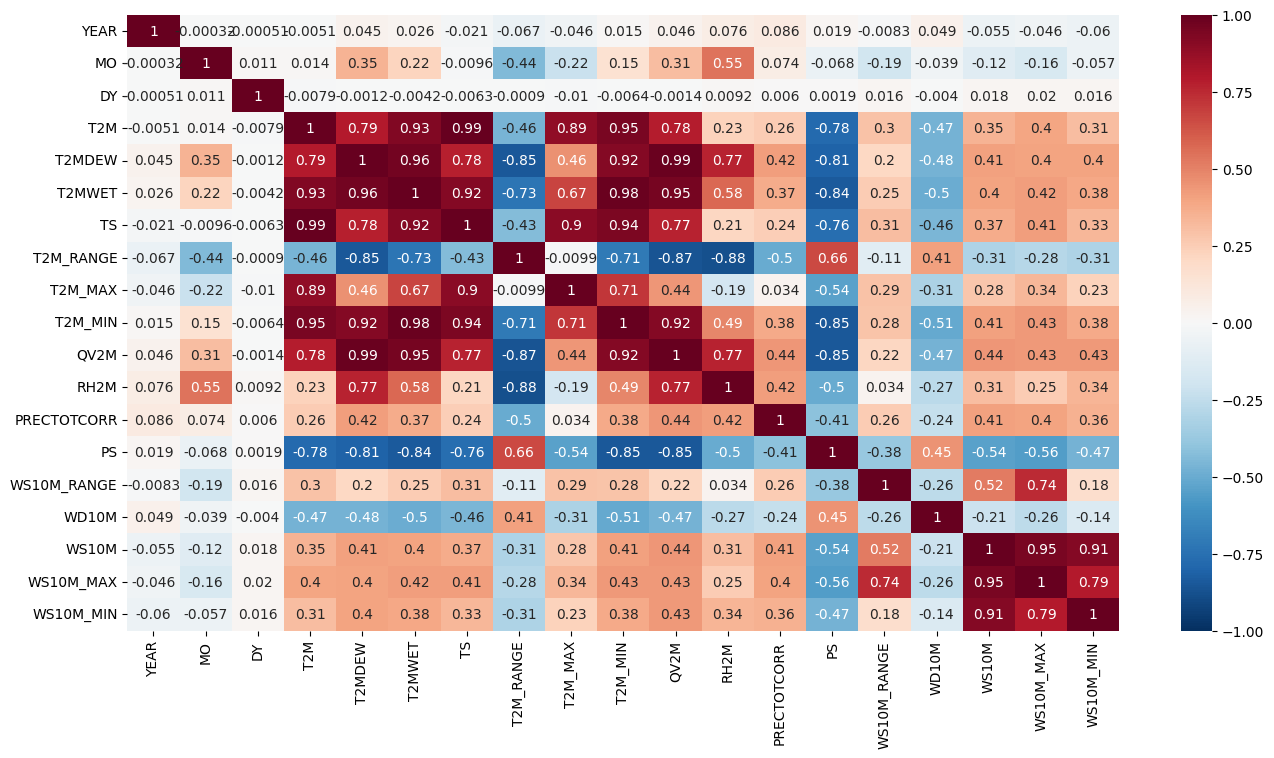}}
\caption{Correlation heatmap among the features of dataset}
\label{fig}
\end{figure}

\begin{figure}[htbp]
\centerline{\includegraphics[width=7cm, height=5cm]{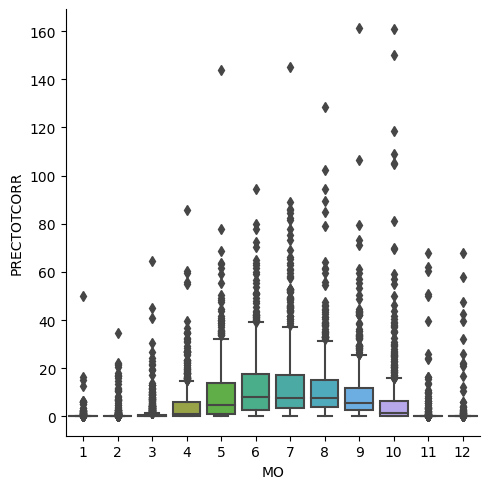}}
\caption{Monthly Precipitation Distribution from the dataset}
\label{fig}
\end{figure}

\begin{figure}[htbp]
\centerline{\includegraphics[width=7cm, height=5cm]{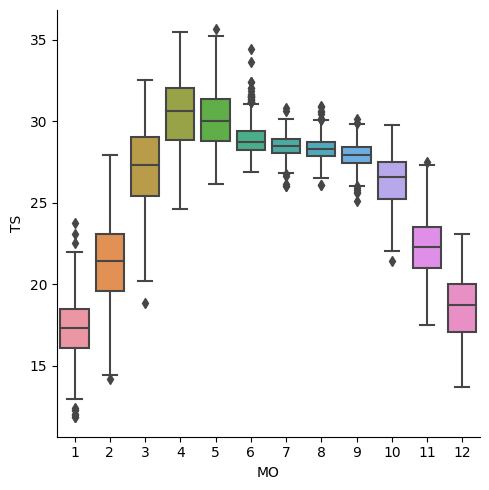}}
\caption{Monthly Temperature Distribution from the dataset}
\label{fig}
\end{figure}

\section{Experiment And Results}

To assess the performance of the proposed weather data prediction model, we employed six different algorithms and evaluated their effectiveness on three target variables. The algorithms are Gradient Boosting, Ada Boost, Artificial Neural Network, Stacking Random Forest, Stacking Neural Network, and Stacking KNN. The dataset was divided into training and testing sets using a random split of 85\% for training and 15\% for testing.

In order to measure the performance of the models, we calculated various evaluation metrics, including accuracy, precision, recall, and F1-score, for each target variable. Higher values of these performance measures indicate better performance of the models. The results presented below demonstrate the performance of the algorithms on the selected target variables, providing valuable insights into their effectiveness in predicting weather data.

\subsection{Precipitation}

The performance measurements of different algorithms for precipitation extraction were analyzed. From table-1, the results showed that all the algorithms performed relatively similar in terms of accuracy, precision, recall, and F1 score. However, the algorithm that achieved the most accurate scores with a minimal difference was Ada Boost. 
Additionally, a confusion matrix was generated to visualize the performance of the best-performing algorithm in predicting precipitation. From the figure-6, in the confusion matrix, the correctly predicted classes are located along the diagonal. This matrix provides an overview of the algorithm's performance in terms of correctly and incorrectly classified instances.

\begin{table}[!t]
\caption{Algorithm Efficiency Overview of Precipitation
}
\begin{center}
$\begin{array}{|c|c|c|c|c|}
\hline \textbf{ Algorithm } & \begin{array}{c}
\textbf{ Accuracy } 
\end{array} & \begin{array}{c}
\textbf { Precision }
\end{array} & \begin{array}{c}
\textbf { Recall } 
\end{array} & \begin{array}{c}
\textbf { F1} \\
\textbf{score}
\end{array} \\
\hline \begin{array}{c}
\text { Gradient  } \\
\text { Boost }
\end{array} & 92.11\% & 89\% & 93\% & 90\%  \\
\hline \begin{array}{c}
\text { Ada} \\
\text { Boost }
\end{array} & 92.51\% & 89\% & 93\% & 90\%\\
\hline \begin{array}{c}
\text { Artificial} \\
\text { Neural }\\
\text{ Network}
\end{array}& 92.15\% & 88\% & 92\% & 89\% \\
\hline \begin{array}{c}
\text { Stacking } \\
\text { Random} \\
\text{Forest}
\end{array}& 92.19\% & 89\% & 92\% & 90\%\\
\hline \begin{array}{c}
\text {Stacking} \\
\text {Neural} \\
\text{Network}
\end{array} & 92.3\% & 89\% & 92\% & 90\%\\
\hline \begin{array}{c}
\text { Stacking } \\
\text { KNN }
\end{array} &91\% & 88\% & 91\% & 89\%\\
\hline 
\end{array}$\\
\end{center}
\end{table}

\begin{figure}[!ht]
\centerline{\includegraphics[width=7cm, height=5cm]{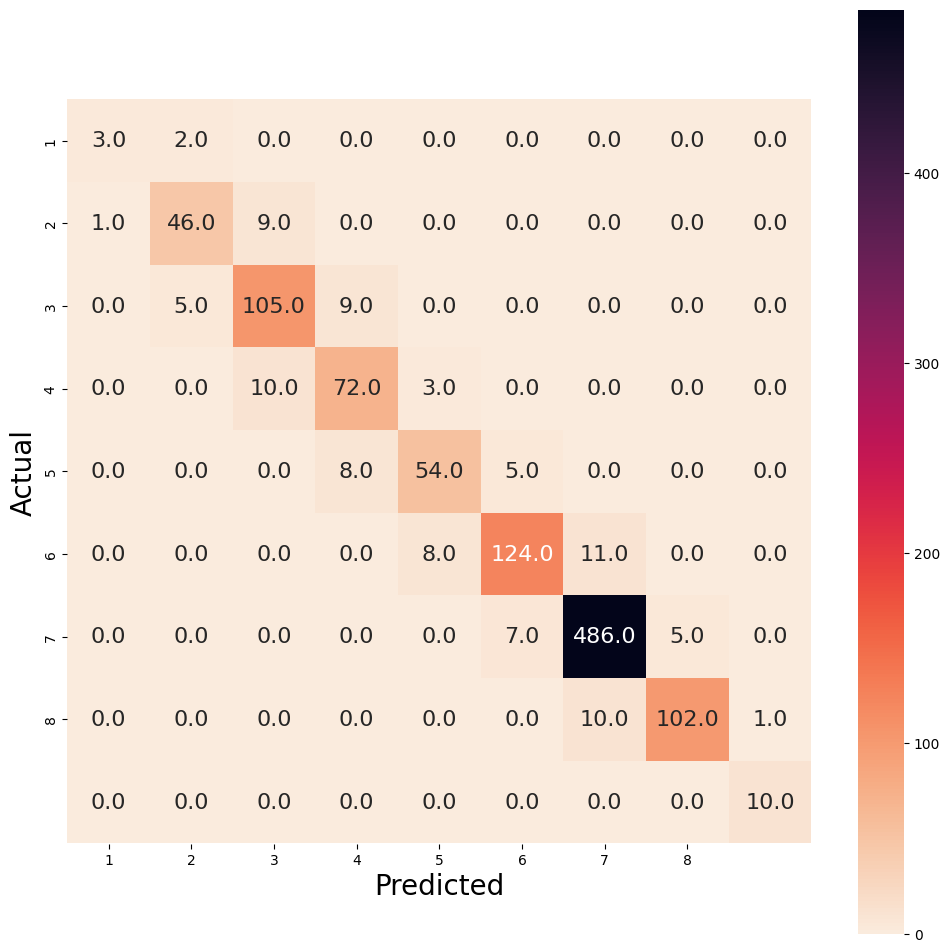}}
\caption{Confusion Matrix for Ada Boost Algorithm (Precipitation)}

\label{fig}
\end{figure}

\begin{table}[htbp]
\caption{Algorithm Efficiency Overview of Temperature
}
\begin{center}
$\begin{array}{|c|c|c|c|c|}
\hline \textbf{ Algorithm } & \begin{array}{c}
\textbf{ Accuracy } 
\end{array} & \begin{array}{c}
\textbf { Precision }
\end{array} & \begin{array}{c}
\textbf { Recall } 
\end{array} & \begin{array}{c}
\textbf { F1} \\
\textbf{score}
\end{array} \\
\hline \begin{array}{c}
\text { Gradient  } \\
\text { Boost }
\end{array} & 91.05\% & 91\% & 91\% & 91\%  \\
\hline \begin{array}{c}
\text { Ada} \\
\text { Boost }
\end{array} & 91.7\% & 92\% & 92\% & 92\%\\
\hline \begin{array}{c}
\text { Artificial} \\
\text { Neural }\\
\text{ Network}
\end{array}& 90.2\% & 90\% & 90\% & 90\% \\
\hline \begin{array}{c}
\text { Stacking } \\
\text { Random} \\
\text{Forest}
\end{array}& 91.3\% & 91\% & 91\% & 91\%\\
\hline \begin{array}{c}
\text {Stacking} \\
\text {Neural} \\
\text{Network}
\end{array} & 91.4\% & 92\% & 91\% & 91\%\\
\hline \begin{array}{c}
\text { Stacking } \\
\text { KNN }
\end{array} &90.96\% & 91\% & 90.6\% & 90.4\%\\
\hline 
\end{array}$\\
\end{center}
\end{table}

\begin{figure}[htbp]
\centerline{\includegraphics[width=7cm, height=5cm]{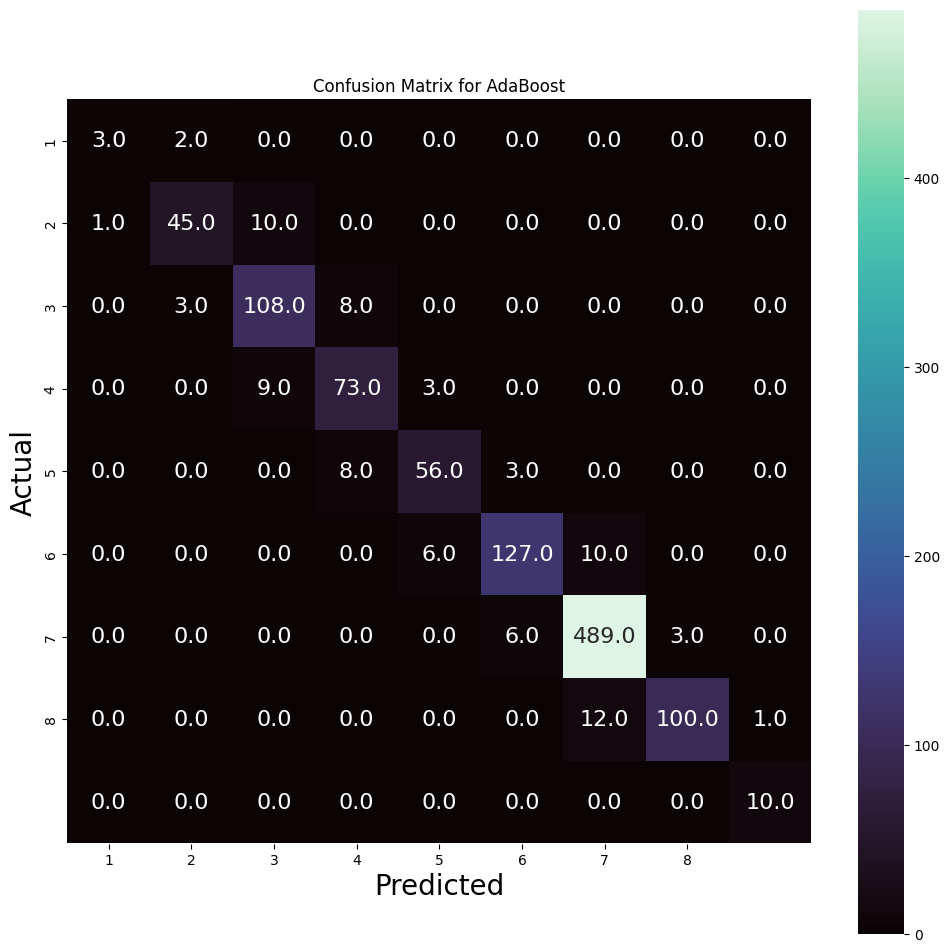}}\caption{Confusion Matrix for Ada Boost Algorithm (Temperature)}
\label{fig}
\end{figure}

\subsection{Temperature}
In the case of average temperature prediction, the same algorithms have shown close pattern results.  On the other hand, the AdaBoost algorithm performed well in this particular task, compared to other algorithms (Table II). These results highlight the superior performance of certain algorithms in accurately predicting average temperature.
Moreover, from the figure-7, in the confusion matrix, the temperature's correctly predicted classes are situated along the diagonal. The provided matrix offers an overview of how the algorithm performed by displaying the count of instances accurately classified and those that were classified incorrectly.

\section{Future Direction}

Our forthcoming research endeavors will center on the integration of advanced models and techniques, with the primary objective of augmenting the precision and dependability of weather forecasting. This will encompass an exploration of ensemble learning methods, deep learning algorithms, and hybrid models that amalgamate multiple approaches. Additionally, considering the substantial data requirements of data-driven methodologies, a new research direction involves efficient feature selection techniques to mitigate computational costs. Furthermore, the incorporation of supplementary meteorological data sources and the exploration of feature engineering techniques to extract more informative features for prediction represent promising avenues of inquiry. The recent domain adaptation for unsupervised learning has a great potential in forecasting. Moreover, the application of big data analytics holds significant potential in effectively managing the vast volume, velocity, and variety of weather data. This entails leveraging distributed computing frameworks such as Apache Hadoop and Apache Spark for processing and storing extensive datasets. Furthermore, the utilization of Internet of Things (IoT) devices and sensor networks facilitate data collection from diverse sources, encompassing weather stations, satellites, drones, buoys, and ground-based sensors. The model's performance evaluation will be bolstered through validation using a wide range of challenging weather datasets.

\section{Conclusion}
In conclusion, the algorithms subjected to testing in this study have exhibited exceptional performance, surpassing conventional weather prediction models. Moreover, the identification of significant weather variables and their corresponding scores has furnished valuable insights into meteorological patterns. Furthermore, our investigation delves into the exploration of influential aspects of weather events, thereby shedding light on critical factors including wind speed at specific altitudes and directions, absolute and relative humidity, precipitation, and surface pressure. Each of these variables is attributed significant scores, underscoring their profound influence on weather patterns and forecasting precision. Collectively, this research emphasizes the potential of the tested algorithms to enhance the accuracy of weather prediction and augment our comprehension of weather phenomena. Consequently, these findings pave the way for advancements in weather forecasting and the formulation of more robust prediction models.

\printbibliography

\vspace{12pt}
\color{red}

\end{document}